\documentclass[10pt,twocolumn,letterpaper]{article}

\usepackage[datasets,pagenumbers]{wacv}

\usepackage{multirow}

\newcommand{\parhead}[1]{%
  \par\smallskip
  \noindent\textbf{#1}%
  \par\vspace{0.1\baselineskip}%
  \noindent
}

\definecolor{wacvblue}{rgb}{0.21,0.49,0.74}
\usepackage[breaklinks,colorlinks,allcolors=wacvblue]{hyperref}

\def\confName{WACV}
\def\confYear{2027}

\title{Visual Dependence Does Not Imply Answerability: A Systematic Analysis of Answerability Signals in Frozen InstructBLIP}

\author{
Minjun Kim\textsuperscript{*}\qquad
Hyunmin Song\textsuperscript{*}\qquad
Changwon Lim\textsuperscript{\textdagger}\\
Chung-Ang University \quad
{\small \{goodwill1669, hmkate, clim\}@cau.ac.kr}\\
{\small \textsuperscript{*}Equal contribution.\quad
\textsuperscript{\textdagger}Corresponding author.}
}

\begin{document}
\maketitle
\begin{abstract}
    Visual question answering models often produce plausible responses even when
    the available visual evidence is insufficient to support a valid answer.
    Although prior work has introduced benchmarks and abstention mechanisms for
    unanswerable visual questions, it remains unclear which signals derived from
    a frozen vision-language model reliably indicate answerability and generalize
    to unseen image--question pairs.
    We systematically evaluate candidate answerability signals on a stratified
    10K subset of UNK-VQA using frozen InstructBLIP, covering predictive
    uncertainty, cross-modal alignment, visual grounding, visual dependence,
    internal representations, and learned probes.
    Across the evaluated methods, answerability discrimination remains weak, with
    test AUROC largely confined to $0.50$--$0.54$.
    CLIP alignment yields the highest point estimate, but does not significantly
    outperform generation confidence under paired bootstrap analysis.
    More strikingly, targeted removal of question-relevant visual evidence causes
    substantially larger changes in answer confidence than same-size random
    masking, yet the resulting intervention sensitivity provides essentially no
    answerability discrimination.
    Increasing probe capacity, combining multiple signals, and adding reason-aware
    supervision do not provide robust improvements, while post-freeze
    cross-validation yields similarly weak discrimination.
    Selective-answering analysis further shows that these weak discrimination
    scores translate into limited usable coverage under low-risk operating
    criteria.
    Failure-mode analysis also shows that the relative utility of individual
    signals varies substantially across perturbation mechanisms.
    Taken together, these results show that, on our UNK-VQA subset with frozen
    InstructBLIP, visual dependence, uncertainty, semantic alignment, and static
    representations are not robust standalone proxies for answerability.
    Our findings emphasize the need to distinguish a model's reliance on visual
    evidence from whether that evidence is sufficient to support an answer.
    \end{abstract}
\section{Introduction}
\label{sec:intro}

Vision-language models (VLMs) have achieved strong performance on visual
question answering (VQA), generating fluent and plausible answers to a wide
range of image--question pairs~\cite{dai2023instructblip,li2023blip2}.
However, the ability to generate a plausible answer does not guarantee that
the answer is supported by sufficient visual evidence.
When an image is ambiguous, relevant information is missing, or a question
cannot be resolved from the available input, a VLM may still produce a
confident response rather than reliably recognizing its
uncertainty~\cite{chen2025unveiling}.
Such overconfident answering limits the reliability of VLMs in settings where
knowing \textit{when not to answer} is as important as generating the answer
itself~\cite{whitehead2022reliable}.

A central challenge is that \textit{confidence} and \textit{answerability}
describe different properties.
Confidence measures how strongly a model favors its prediction, whereas
answerability concerns whether the available image--question pair contains
sufficient evidence to support a valid answer.
A model can therefore be confident even when the evidence needed to justify
its answer is incomplete, ambiguous, or unavailable.
Reliable answerability estimation consequently requires understanding whether
the available evidence is sufficient, rather than relying on prediction
confidence alone.

This distinction motivates us to examine \textit{visual dependence} as a
candidate answerability signal.
If a question is answerable from an image, the model's prediction may depend
on question-relevant visual evidence, and removing that evidence should
produce a measurable change in model behavior.
This leads to a natural hypothesis: stronger dependence on question-relevant
visual evidence may indicate that an image--question pair is answerable.
However, visual dependence and evidence sufficiency need not be equivalent.
A model may rely strongly on the visual information that is available while
still lacking other evidence required for a valid answer.
Whether visual dependence provides a reliable proxy for answerability is
therefore an empirical question.

Rather than treating answerability detection as another model-design problem,
we treat candidate answerability proxies themselves as objects of evaluation.
Our goal is to audit whether commonly used model-side signals support the
evaluative claim that an image--question pair is answerable, and whether such
conclusions remain stable across controlled perturbations, resampling, and
heterogeneous failure modes.

We investigate this question on UNK-VQA~\cite{guo2024unkvqa}, a benchmark
containing both answerable and unanswerable image--question pairs, using a
common frozen-VLM evaluation framework.
Beyond visual dependence, we examine predictive uncertainty, cross-modal
alignment, visual grounding, and internal representations, together with
feature transformation, feature fusion, nonlinear probing, and reason-aware
supervision.
Our study addresses three main questions:
\textit{(1) How reliably do uncertainty, cross-modal alignment, visual
grounding, visual dependence, and internal representations distinguish
answerable from unanswerable inputs?
(2) Can representation transformation, feature fusion, increased probe
capacity, or reason-aware supervision improve answerability generalization?
(3) Do these signals remain reliable across different perturbation mechanisms
and semantic reasons for unanswerability?}
We additionally evaluate statistical robustness and selective-answering
behavior to determine whether the observed discrimination translates into
reliable operating behavior.

Our experiments reveal consistently weak answerability discrimination.
Across the evaluated signals and probes, test AUROC remains approximately
0.50--0.54.
Targeted removal of question-relevant visual evidence affects model confidence
substantially more than same-size random removal (0.1426 vs.\ 0.0258), yet the
resulting intervention sensitivity remains near chance for answerability
discrimination (AUROC 0.5004).
Representation transformation, feature fusion, nonlinear probing,
reason-aware supervision, and selective-answering analysis do not yield robust
or practically useful answerability behavior, and failure-mode analysis shows
that signal utility varies across forms of unanswerability.

Our contributions are threefold.
First, we conduct a systematic validity audit of candidate answerability
proxies within a common frozen-VLM framework, spanning predictive uncertainty,
cross-modal alignment, visual grounding, visual dependence, internal
representations, feature fusion, nonlinear probing, and reason-aware
supervision.
Second, through targeted evidence removal, we empirically show that measurable
visual dependence does not translate into answerability: intervention
sensitivity remains near chance for answerable--unanswerable discrimination,
separating model reliance from evidence sufficiency.
Third, we conduct a generalization and failure-mode audit using paired
bootstrap analysis, multiple seeds, post-freeze cross-validation, and
perturbation- and reason-wise evaluation, showing that no evaluated signal or
probe provides consistently robust discrimination across heterogeneous forms of
unanswerability.

Together, these findings suggest that reliable answerability estimation
requires reasoning about whether the evidence required by a question is
actually available and sufficient, rather than relying on model confidence,
visual dependence, or static representation separability alone.

\section{Related Work}
\label{sec:related}

\subsection{Reliable VQA and Selective Prediction}

Visual Question Answering (VQA) requires models to answer natural-language
questions based on visual content~\cite{antol2015vqa,goyal2017vqav2}, and
recent VLMs have improved VQA by integrating pretrained visual and language
representations~\cite{dai2023instructblip,li2023blip2}.
However, strong answer generation does not guarantee reliable answering:
models may produce plausible responses that are not visually supported,
motivating work on hallucination, selective prediction, and abstention
~\cite{li2023pope,geifman2017selective,whitehead2022reliable,dancette2023improving,srinivasan2024selective}.
These approaches primarily concern prediction reliability, whereas
answerability concerns whether the available evidence supports a valid answer.

\subsection{Unanswerable VQA and UNK-VQA}

Unanswerable VQA explicitly considers cases in which an image--question pair
does not provide sufficient evidence for a valid
answer~\cite{gurari2018vizwiz,miyai2025unsolvable}.
UNK-VQA~\cite{guo2024unkvqa} provides a controlled benchmark with answerable
and unanswerable image--question pairs annotated by different causes of
unanswerability.
Recent studies further show that such refusal behavior is strongly influenced
by the available visual input and remains challenging for current multimodal
models~\cite{zhu2026mohobench}.
Concurrent work also separates visual modality dependency from evidence
sufficiency when constructing unanswerable
instances~\cite{madhusudhan2026knowing}.
Whereas that work uses the distinction to build a benchmark and evaluate
prompted abstention behavior, we test the distinction directly at the signal
level: we measure whether an intervention-based dependence score, computed from
a frozen VLM, actually discriminates answerability.

Our work builds on this setting but focuses on the signals underlying
answerability decisions.
Rather than proposing a new abstention mechanism, we systematically examine
whether signals available from a frozen VLM can distinguish answerable from
unanswerable inputs across different forms of unanswerability.

\subsection{Answerability Signals and Evidence Sufficiency}

Predictive uncertainty provides a natural signal for estimating whether a
model should answer~\cite{hendrycks2017baseline}.
Generation confidence reflects the probability assigned to a generated
response, while predictive entropy characterizes uncertainty in the predictive
distribution~\cite{malinin2021uncertainty}.
Such signals are commonly used for selective prediction, but they characterize
uncertainty in model predictions rather than directly measuring the
sufficiency of the available evidence.

Cross-modal alignment provides another potential signal.
CLIP-based similarity has been used to measure semantic compatibility between
visual and textual content~\cite{hessel2021clipscore,radford2021clip}.
More directly, CLIP-UP~\cite{vardi2026clipup} combines CLIP-derived
image--question alignment with lightweight learned layers on a frozen VLM to
detect unanswerable questions.
In contrast, we examine whether cross-modal alignment itself serves as a robust
standalone proxy for answerability across heterogeneous controlled
perturbations.
Visual grounding provides a related perspective by localizing image regions
associated with textual concepts~\cite{liu2024groundingdino}.
Alignment and grounding therefore characterize the relationship between the
question and visual content, but their relationship to answerability and
evidence sufficiency remains unclear; prior work has questioned whether
improvements associated with visual grounding necessarily reflect stronger use
of the correct visual evidence~\cite{shrestha2020negative}.

Counterfactual evidence intervention provides a more direct way to examine
visual dependence.
By removing or modifying question-relevant visual evidence and measuring the
resulting change in model behavior, intervention-based analysis can test
whether a prediction depends on particular image
regions~\cite{petsiuk2018rise,leng2024vcd}.
EVICT~\cite{kotte2026evict} applies counterfactual evidence dropout for
selective VQA by masking cited visual evidence and measuring the resulting
answer change, reporting improvements in some visually grounded QA settings but
weaker behavior on hard-negative cases.
Related analyses similarly suggest that apparent reliance on visual content
does not necessarily imply reliable visual understanding or answer
correctness~\cite{chen2024mmstar,liu2025seeing}.
This motivates a distinction between \textit{visual dependence} and
\textit{evidence sufficiency}: the former concerns how model behavior changes
when relevant visual evidence is altered, whereas the latter concerns whether
the available evidence is adequate to support a valid answer.
We therefore ask whether intervention sensitivity itself generalizes as a
robust proxy for answerability across heterogeneous unanswerability
mechanisms.

Frozen multimodal representations may also contain answerability information
that is not directly captured by uncertainty, alignment, grounding, or
intervention scores.
Probing these representations can test whether answerable and unanswerable
inputs are separable in the learned feature
space~\cite{alain2017understanding}.
Recent analyses of language models nonetheless caution that internal states may
reflect knowledge recall rather than a reliable notion of what a model does not
know~\cite{cheang2026llms}.
Feature combinations and nonlinear probes further provide a way to test
whether answerability information can be recovered from multiple signals.

Overall, these approaches capture different aspects of model uncertainty,
cross-modal compatibility, visual grounding, evidence sensitivity, and
representation structure.
However, it remains unclear whether any of these signals provides a reliable
proxy for evidence sufficiency under a common evaluation setting.
Our work addresses this question through a systematic evaluation of
answerability signals extracted from a frozen VLM.

\section{Evaluation Framework}
\label{sec:framework}

\subsection{UNK-VQA Benchmark}
\label{subsec:unkvqa}

UNK-VQA~\cite{guo2024unkvqa} applies controlled perturbations to VQA
v2~\cite{goyal2017vqav2} and annotates the resulting image--question pairs for
answerability. Its five perturbations are word replacement (T-1), semantic
negation (T-2), image replacement (I-1), image masking (I-2), and copy-and-move
(I-3). Unanswerable examples are further labeled with one of four official
UNK-VQA reasons: it has multiple plausible answers (R1),
it is difficult to understand (R2),
the image lacks important concepts/information (R3), or
it requires higher-level knowledge to answer (R4); both taxonomies are analyzed
in Sec.~\ref{sec:failure}.

\parhead{Experimental subset construction.}The full release contains 99,996 examples and is imbalanced in answerability
and perturbation type. To compare heterogeneous failure modes, we draw a
stratified 10K subset from the released splits, balancing answerability and
perturbation type within 7,000/1,000/2,000 train/validation/test examples
(Table~\ref{tab:dataset_stats}).

\begin{table*}[t]
    \centering
    \footnotesize
    \setlength{\tabcolsep}{5pt}
    \begin{tabular}{lrc|ccccc}
        \toprule
        Split & Total & A\,/\,U
            & T-1 & T-2 & I-1 & I-2 & I-3 \\
        \midrule
        Train      & 7,000  & 3,500\,/\,3,500
            & 700\,/\,700 & 700\,/\,700 & 700\,/\,700 & 700\,/\,700 & 700\,/\,700 \\
        Validation & 1,000  & 500\,/\,500
            & 100\,/\,100 & 100\,/\,100 & 100\,/\,100 & 100\,/\,100 & 100\,/\,100 \\
        Test       & 2,000  & 1,000\,/\,1,000
            & 200\,/\,200 & 200\,/\,200 & 200\,/\,200 & 200\,/\,200 & 200\,/\,200 \\
        \midrule
        Total      & 10,000 & 5,000\,/\,5,000
            & 1,000\,/\,1,000 & 1,000\,/\,1,000 & 1,000\,/\,1,000 & 1,000\,/\,1,000 & 1,000\,/\,1,000 \\
        \bottomrule
    \end{tabular}
    \caption{Statistics of the stratified UNK-VQA subset used in our
    experiments.
    Each split is balanced across binary answerability (A/U) and the five
    perturbation types; cell entries report answerable/unanswerable counts.}
    \label{tab:dataset_stats}
\end{table*}

We enforce source-question disjointness across splits using the dataset's
\texttt{question\_id\_origin} field.
Reasons are not stratified, preserving their natural imbalance: among the
5,000 unanswerable examples, R1--R4 account for 19.2\%, 34.9\%, 22.6\%, and
23.4\%, respectively.

\subsection{Frozen VLM and Answerability Signals}
\label{subsec:signals}

We keep InstructBLIP~\cite{dai2023instructblip} with
Flan-T5-XL~\cite{chung2022flant5} fully frozen and test signals from its
outputs, visual evidence, and internal representations.

\parhead{Predictive uncertainty.}Generation confidence is the geometric mean of the probabilities of
greedily selected answer tokens, excluding special tokens.
Confidence (1-D LR) is an L2-regularized one-dimensional logistic-regression
probe on this scalar; we also use negative predictive entropy, so larger scores
indicate answerability.

\parhead{Cross-modal alignment.}We use CLIP ViT-B/32~\cite{radford2021clip} image--question cosine similarity
directly as an alignment score.

\parhead{Visual grounding.}Grounding DINO~\cite{liu2024groundingdino} and
SAM~\cite{kirillov2023sam} localize question-relevant regions.
\textit{Grounding max} is the maximum grounding score; \textit{Grounding
features (3-D)} adds required-entity coverage and mask-area ratio.
Grounding features are computed from Question-Query detections only; the
optional Answer Query is used solely to refine the intervention target when the
question alone does not identify a stable object-level target.

\parhead{Visual dependence under evidence intervention.}We remove question-relevant visual evidence and measure the resulting change in
model confidence.
We first analyze each question to identify its question type, required visual
entities, relations, and potential target, and determine whether the visual
target can be specified from the question alone.
Based on this analysis, we construct either a Question Query from directly
identifiable visual entities or, only when the target cannot be sufficiently
specified from the question, an Answer Query that uses InstructBLIP's original
generated answer to refine the target.
Dataset ground-truth answers are never used.

Before grounding, each example is assigned one of three mask strategies:
question-primary, when the target can be identified from the question alone;
answer-primary, when the original generated answer is required to refine the
target; or none, when no stable object-level target can be identified.
This strategy is fixed before Grounding DINO~\cite{liu2024groundingdino}
detection rather than selected post hoc from grounding or intervention outcomes.

Grounding DINO localizes the query-relevant object, and the selected bounding
box is refined into a pixel-level mask using SAM~\cite{kirillov2023sam}.
The masked region is replaced with the image mean pixel value to construct the
evidence-removed image $I_{\mathrm{rem}}$.
When no valid target mask is obtained, a zero mask is used and the example
remains in the evaluation.

Let $A_{\mathrm{orig}}$ denote the answer generated from the original image $I$
and question $q$.
We keep $A_{\mathrm{orig}}$ fixed and define \textit{intervention sensitivity}
as

\begin{equation}
    s_{\mathrm{int}}
    =
    c(A_{\mathrm{orig}} \mid I,q)
    -
    c(A_{\mathrm{orig}} \mid I_{\mathrm{rem}},q),
\end{equation}
\noindent where larger values indicate stronger visual dependence.
We denote this score \textit{Intervention}.
A same-size random mask is used as a control for generic sensitivity to image
removal~\cite{petsiuk2018rise}.
For the multi-signal probe, we separately regenerate an answer from
$I_{\mathrm{rem}}$; these consistency features are distinct from fixed-answer
rescoring.

\parhead{Internal representations.}We also probe frozen Q-Former~\cite{li2023blip2} and question representations,
their PCA reductions, and their combinations with other signals.

Table~\ref{tab:signals} summarizes the answerability signals and probe
configurations evaluated in our framework.

\begin{table}[t]
    \centering
    \footnotesize
    \setlength{\tabcolsep}{2pt}
    \begin{tabular}{@{}lll@{}}
        \toprule
        Signal type & Method & Score \\
        \midrule
        \multirow{2}{*}{\shortstack[l]{Predictive\\uncertainty}}
            & Confidence (1-D LR) & 1-D LR prob. \\
            & Entropy & Neg.\ entropy \\
        \midrule
        \shortstack[l]{Cross-modal\\alignment}
            & CLIP Alignment & Cosine sim. \\
        \midrule
        Visual grounding
            & Grounding max & Max grounding \\
        Visual dependence
            & Intervention & Conf.\ drop \\
        \midrule
        Multi-signal
            & 7-feature LR & LR score \\
        \midrule
        \multirow{2}{*}{\shortstack[l]{Internal\\representation}}
            & Q-Former & \multirow{2}{*}{Probe score} \\
            & Question & \\
        \midrule
        \multirow{3}{*}{\shortstack[l]{Feature\\fusion}}
            & Q-Former + Question & \multirow{3}{*}{LR score} \\
            & + Confidence & \\
            & + Grounding features (3-D) & \\
        \midrule
        Nonlinear probe
            & Binary MLP & \multirow{2}{*}{Probe score} \\
        Reason-aware probe
            & Reason-aware MTL & \\
        \bottomrule
    \end{tabular}
    \caption{Answerability signals and probe configurations evaluated in our
    framework.}
    \label{tab:signals}
    \vspace{-8pt}
\end{table}

\subsection{Answerability Probes}
\label{subsec:probes}

We use L2-regularized logistic regression unless otherwise stated to test
whether individual or combined frozen features encode
answerability~\cite{alain2017understanding}.

\parhead{Multi-signal probe.}The \textit{7-feature LR} combines generation confidence, fixed-answer
confidence drop, exact and semantic consistency between the original and
removal-generated answers, maximum grounding score, required-entity coverage,
and mask-area ratio. The consistency features use generated answers, never
dataset ground truth, and are distinct from the confidence drop obtained by
re-scoring the fixed original answer.

\parhead{Representation and feature-fusion probes.}We probe Q-Former and question representations individually and jointly,
incrementally add confidence and Grounding features (3-D), and test PCA reduction.

\parhead{Nonlinear and reason-aware probes.}We also test a binary MLP~\cite{hewitt2019designing} and a reason-aware MTL
probe~\cite{ruder2017overview} that jointly predicts answerability and
unanswerability reason. All configurations are selected on validation data.

\subsection{Evaluation Protocol}
\label{subsec:evaluation}

We treat answerable examples as positive and primarily report AUROC, where 0.5
is random ranking. All parameters, scaling, and PCA are fitted on training data;
hyperparameters are selected on validation data.

Feature extraction is complete for validation and test; all primary analyses
use the complete 2,000-example test subset.

\parhead{Threshold-dependent evaluation.}One global validation threshold maximizing balanced accuracy is applied
unchanged to the full test set and perturbation subsets for unanswerable recall
(U-Recall).

\parhead{Statistical evaluation.}We estimate 95\% bootstrap confidence intervals for test AUROC and paired
bootstrap differences relative to the Confidence baseline.
Nonlinear probes are additionally evaluated across three random seeds.
Because the experimental test subset was inspected during analysis development,
we perform a post-freeze 5-fold cross-validation robustness check on the
training subset after fixing all evaluated methods and hyperparameters; these
results are not used for additional model or hyperparameter selection.

\parhead{Failure-mode evaluation.}We report perturbation-wise AUROC and U-Recall and reason-wise AUROC.
Reason-wise AUROC compares each unanswerable group with the common 1,000
answerable examples.

\parhead{Selective answering.}Test examples are ranked in descending order of answerability score.
For the top-$k$ retained subset $S_k$, we define coverage as $C(k)=k/N$ and
selective risk as the fraction of unanswerable examples among retained
examples,
\begin{equation}
    R(k)=\frac{1}{k}\sum_{i \in S_k}\mathbb{1}(y_i=U).
\end{equation}
We evaluate all $k=1,\ldots,N$, report area under the risk--coverage curve
(AURC; lower is better), and report the maximum coverage satisfying
$R(k)\leq0.3$~\cite{geifman2019bias,traub2024overcoming}.

\section{Systematic Evaluation}
\label{sec:systematic}

We evaluate whether signals available from a frozen VLM can reliably
distinguish answerable from unanswerable image--question pairs.
We first compare candidate answerability signals under a common protocol,
then examine visual dependence, representation and probe capacity, statistical
robustness, and selective-answering behavior.

\subsection{Overall Answerability Signal Benchmark}
\label{subsec:overall}

\begin{figure}[t]
    \centering
    \includegraphics[width=\linewidth]{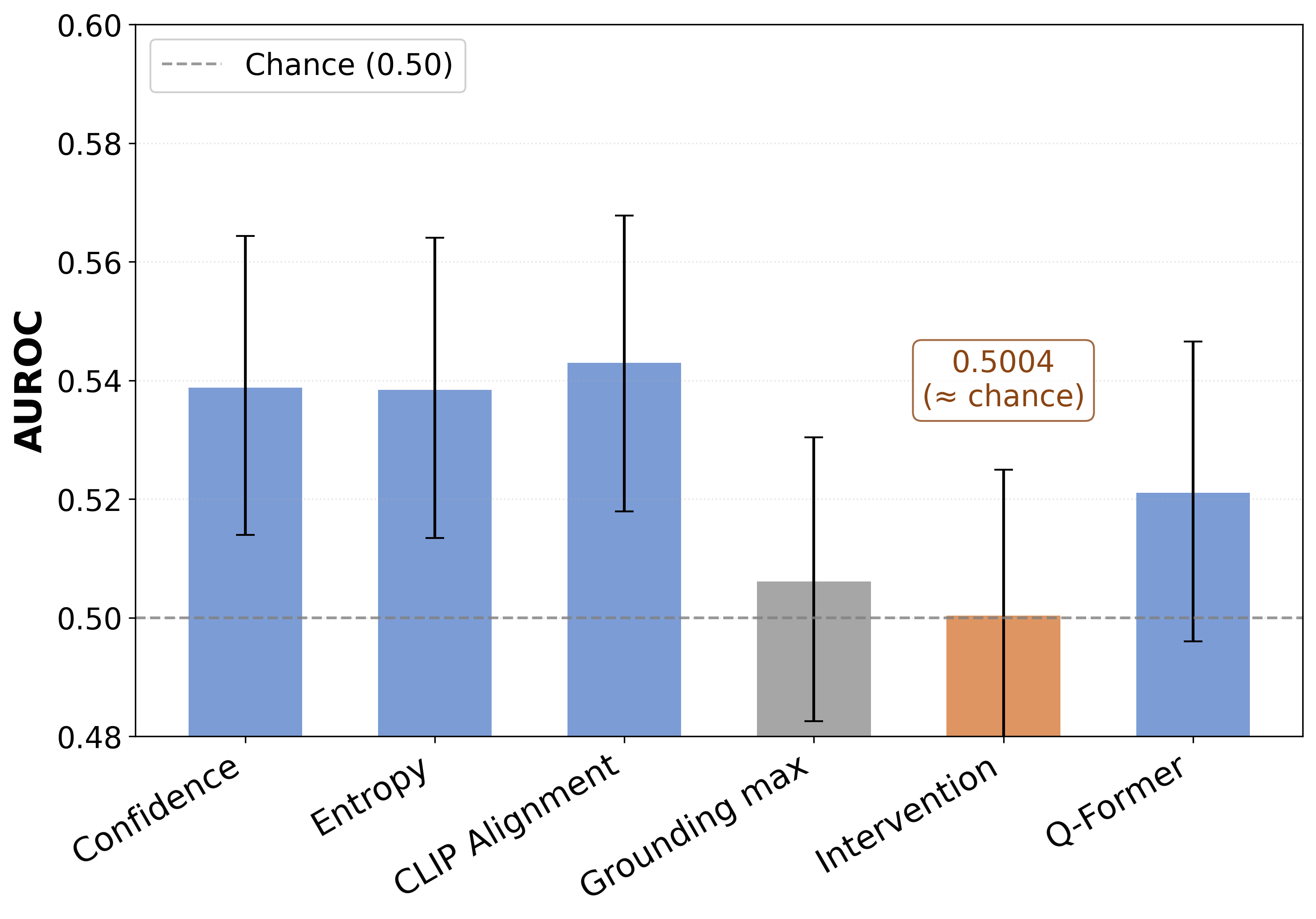}
    \caption{Test AUROC with 95\% bootstrap confidence intervals for
    representative standalone signals on the 2,000-example test subset.
    The dashed line marks chance (0.50).}
    \label{fig:auroc_ci}
\end{figure}

Table~\ref{tab:overall_results} summarizes the discriminative ability of the
candidate answerability proxies on the complete 2,000-example experimental test
subset.
Fig.~\ref{fig:auroc_ci} reports bootstrap confidence intervals for
representative standalone signals.
Across all evaluated signal families, performance remains close to random
ranking, with AUROC ranging from 0.5004 to 0.5430.
CLIP Alignment yields the highest point estimate (0.5430), followed closely by
the Confidence baseline (0.5388) and Entropy (0.5384).
In contrast, Grounding max (0.5061) and Intervention (0.5004) provide little
standalone discrimination.

\begin{table}[t]
    \centering
    \footnotesize
    \setlength{\tabcolsep}{3pt}
    \begin{tabular}{llc}
        \toprule
        Signal type & Method & Test AUROC \\
        \midrule
        \multirow{2}{*}{\shortstack[l]{Predictive\\uncertainty}}
            & Confidence (1-D LR) & 0.5388 \\
            & Entropy & 0.5384 \\
        \midrule
        \shortstack[l]{Cross-modal\\alignment}
            & CLIP Alignment & \textbf{0.5430} \\
        \midrule
        Visual grounding
            & Grounding max & 0.5061 \\
        Visual dependence
            & Intervention & 0.5004 \\
        \midrule
        Multi-signal
            & 7-feature LR & 0.5296 \\
        \midrule
        \multirow{2}{*}{\shortstack[l]{Internal\\representation}}
            & Q-Former & 0.5210 \\
            & Question & 0.5111 \\
        \midrule
        \multirow{3}{*}{\shortstack[l]{Feature\\fusion}}
            & Q-Former + Question & 0.5293 \\
            & + Confidence & 0.5300 \\
            & + Grounding features (3-D) & 0.5300 \\
        \midrule
        Nonlinear probe
            & Binary MLP & 0.5147 \\
        Reason-aware probe
            & Reason-aware MTL & 0.5204 \\
        \bottomrule
    \end{tabular}
    \caption{Overall answerability discrimination on the complete
    2,000-example experimental test subset.}
    \label{tab:overall_results}
    \vspace{-8pt}
\end{table}

Combining heterogeneous signals does not recover substantially stronger
answerability information.
The 7-feature LR obtains 0.5296 AUROC, while combining Q-Former and Question
representations reaches 0.5293; adding Confidence and Grounding features
increases performance only marginally to 0.5300.
Increasing probe capacity also fails to improve generalization, with Binary
MLP and reason-aware MTL obtaining 0.5147 and 0.5204 AUROC, respectively.

Taken together, these results provide little evidence that any single evaluated
proxy, representation, or lightweight combination offers robust standalone
answerability discrimination in this frozen-InstructBLIP setting.
We examine the statistical support for the small differences among the
strongest methods in Sec.~\ref{subsec:reliability}.

\subsection{Visual Dependence Does Not Imply Answerability}
\label{subsec:intervention}

We next examine whether sensitivity to question-relevant visual evidence
provides a useful proxy for answerability.
Under the fixed-answer intervention protocol in Sec.~\ref{subsec:signals}, we
apply the same pipeline to the test subset under two masking conditions:
question-relevant \textit{targeted} removal and a same-size \textit{random}
mask control.
In each run, we keep the original generated answer fixed and measure the
resulting confidence drop.

\begin{figure}[t]
    \centering
    \includegraphics[width=\linewidth]{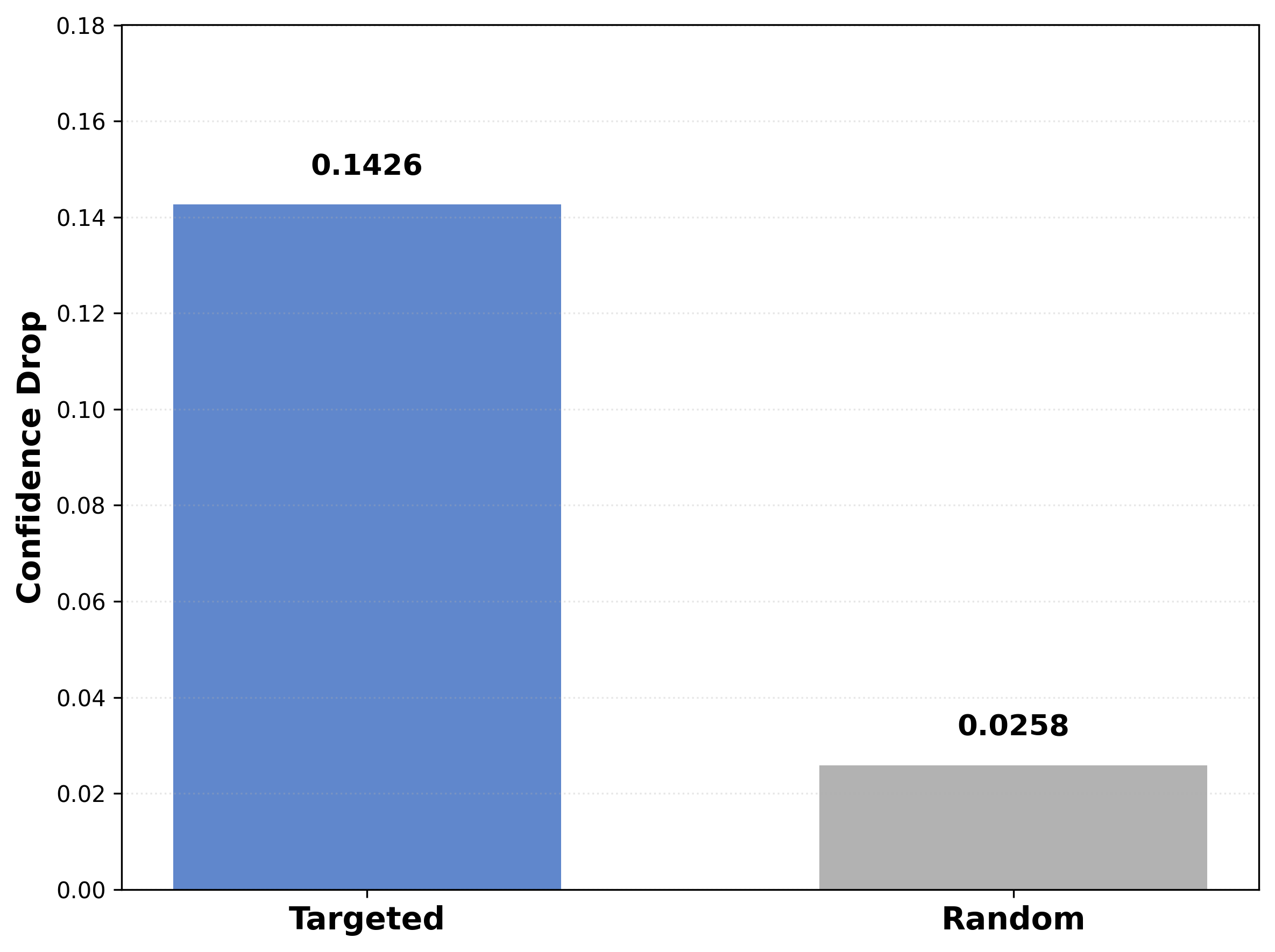}
    \caption{Mean fixed-answer confidence drop under targeted versus same-size
    random masking on the 2,000-example test subset.
    Both bars use the same intervention protocol; only the masking condition
    differs.}
    \label{fig:confidence_drop}
    \vspace{-8pt}
\end{figure}

Fig.~\ref{fig:confidence_drop} compares the two conditions.
When evidence is removed with targeted masking, the mean fixed-answer
confidence drop is 0.1426; under the random control, it is 0.0258.
The frozen VLM is therefore measurably more sensitive to removal of
question-relevant regions than to arbitrary regions of the same size, indicating
that the intervention captures a form of visual dependence.

\begin{figure*}[t]
    \centering
    \includegraphics[width=\textwidth]{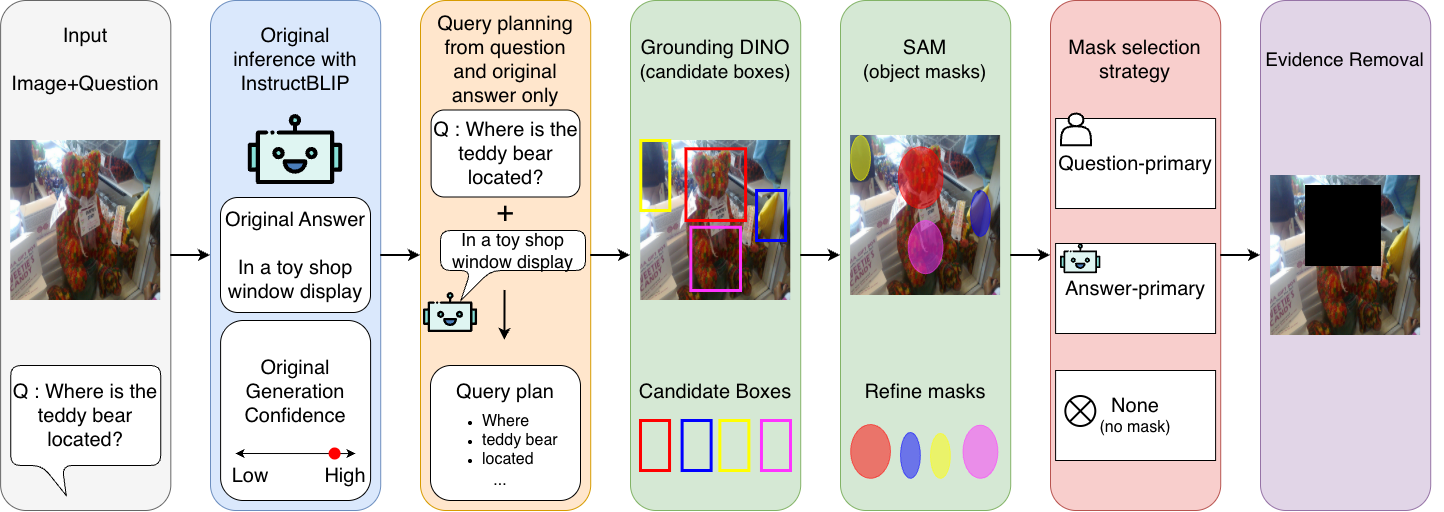}
    \caption{Full evidence-intervention pipeline shared by both masking
    conditions (Sec.~\ref{subsec:signals}).
    InstructBLIP generates a fixed original answer and confidence score from
    the unmodified image.
    Each example is assigned a question-primary, answer-primary, or no-mask
    strategy before detection; Grounding DINO~\cite{liu2024groundingdino} and
    SAM~\cite{kirillov2023sam} localize the query-relevant region, which is
    removed via mean-pixel replacement to form the evidence-removed image
    $I_{\mathrm{rem}}$.
    The same fixed answer is re-scored on $I_{\mathrm{rem}}$.
    Dataset ground-truth answers are never used.}
    \label{fig:intervention}
\end{figure*}

Fig.~\ref{fig:intervention} summarizes the full procedure used for both runs.
Crucially, however, this dependence does not translate into answerability
discrimination: using the targeted fixed-answer confidence drop as an
answerability score yields AUROC 0.5004, effectively corresponding to random
ranking.
Both answerable and unanswerable examples can exhibit strong sensitivity to
removal of question-relevant visual evidence.

A valid target mask is obtained for 54.7\% of the experimental test subset.
Under our predefined protocol, invalid localizations are assigned zero masks and
remain in the evaluation; therefore, the reported AUROC over all 2,000 test
examples measures the end-to-end discriminative utility of the complete
intervention pipeline, including localization failure.

Thus, within the evaluated intervention pipeline, measurable visual dependence
is not a reliable standalone proxy for answerability.

\subsection{Representation and Learned-Probe Analysis}
\label{subsec:learned}

We next test whether the weak discrimination can be improved through
representation transformation, feature fusion, increased nonlinear capacity,
or explicit supervision over reasons for unanswerability.

\parhead{Representation reduction and feature fusion.}PCA provides no meaningful improvement, and feature fusion improves only
marginally over individual frozen representations.
These results suggest that the weak discrimination is not explained simply by
high representation dimensionality or the absence of a particular signal.

\parhead{Nonlinear and reason-aware probes.}The binary MLP reaches training, validation, and test AUROCs of 0.7718,
0.5585, and 0.5147, respectively.
The large train--test gap is consistent with substantial overfitting rather
than improved generalizable answerability discrimination.

Reason-aware MTL similarly provides no clear advantage, while its auxiliary
reason predictions are strongly concentrated on Unclear/Difficult (R2), indicating poor separation
among unanswerability reasons.

Overall, dimensionality reduction, feature fusion, nonlinear capacity, and
reason-aware supervision do not resolve the weak generalization of the
evaluated frozen-VLM signals.

\subsection{Statistical Robustness}
\label{subsec:reliability}

Because the differences among the strongest methods are small, we examine
whether the conclusions persist under bootstrap resampling, random
initialization, and alternative training partitions.

\parhead{Paired bootstrap.}CLIP Alignment has the highest point estimate, but its AUROC difference
relative to Confidence is only $+0.0042$, with a 95\% paired bootstrap
confidence interval of $[-0.0298, 0.0377]$.
Thus, the observed improvement over Confidence is not statistically supported.
The paired intervals for the other evaluated alternatives also include zero,
except for Intervention, which is lower than Confidence by $-0.0384$
(95\% CI $[-0.0698,-0.0061]$).

\parhead{Multiple seeds and post-freeze cross-validation.}Across seeds 42, 43, and 44, Binary MLP obtains
$0.5153 \pm 0.0059$ test AUROC and reason-aware MTL obtains
$0.5153 \pm 0.0050$, providing no evidence of a seed-dependent improvement.

After freezing the evaluated methods and hyperparameters, a post-freeze 5-fold
cross-validation robustness check on the training subset yields mean AUROCs from
0.5144 to 0.5290.
The similarly weak cross-validation results support the conclusion that the
observed limitation is not specific to a single experimental partition.

\subsection{Selective Answering under Risk--Coverage}
\label{subsec:selective}

Finally, we examine whether the strongest simple answerability scores provide
useful operating points for selective answering~\cite{srinivasan2024selective}.
Entropy, Confidence, and CLIP Alignment obtain AURCs of 0.4577, 0.4626, and
0.4641, respectively.
At risk $\leq 0.3$, Confidence retains only 2.85\% of the test examples,
Entropy retains 0.20\%, and CLIP Alignment provides no feasible operating
point.

Thus, the weak AUROC discrimination also translates into limited practical
selective-answering behavior: the evaluated simple signals cannot identify a
substantial subset of examples that can be retained under the specified
low-risk criterion.

\section{Failure-Mode Analysis}
\label{sec:failure}

We evaluate five representative methods---Confidence, CLIP Alignment,
Intervention, Q-Former, and Binary MLP---across perturbation and reason
categories using the same frozen scores and validation-selected thresholds;
no model or threshold is fitted to a test subgroup.

\subsection{Perturbation-Wise Performance}
\label{subsec:perturbation}

Table~\ref{tab:perturbation_failure} reports AUROC and U-Recall for each
method and perturbation type.

\begin{table*}[t]
    \centering
    \footnotesize
    \setlength{\tabcolsep}{4pt}
    \begin{tabular}{lccccc}
        \toprule
        Method & T-1 & T-2 & I-1 & I-2 & I-3 \\
        \midrule
        Confidence
        & 0.521/0.275 & 0.579/0.555 & 0.601/0.395 & 0.478/0.285 & 0.517/0.245 \\

        CLIP Alignment
        & 0.561/0.725 & 0.539/0.590 & 0.550/0.890 & 0.498/0.520 & 0.571/0.655 \\

        Intervention
        & 0.488/0.780 & 0.487/0.850 & 0.528/0.835 & 0.482/0.795 & 0.519/0.745 \\

        Q-Former
        & 0.470/0.210 & 0.553/0.250 & 0.555/0.290 & 0.510/0.430 & 0.519/0.295 \\

        Binary MLP
        & 0.527/0.345 & 0.495/0.315 & 0.556/0.405 & 0.508/0.450 & 0.485/0.390 \\
        \bottomrule
    \end{tabular}
    \caption{Perturbation-wise answerability performance.
    Each entry reports AUROC / U-Recall.
    U-Recall is computed using the single global threshold selected on the
    validation subset; no threshold is re-selected for a perturbation type.}
    \label{tab:perturbation_failure}
\end{table*}

No method dominates all perturbations, and all are near chance on I-2.
Intervention combines high fixed-threshold U-Recall with near-chance AUROC,
showing that rejecting many unanswerable cases at one operating point does not
imply reliable ranking.

\subsection{Reason-Wise Performance}
\label{subsec:reason}

Reason groups retain their natural imbalance; each is compared with the common
set of 1,000 answerable test examples.

\begin{table}[t]
    \centering
    \footnotesize
    \setlength{\tabcolsep}{3pt}
    \begin{tabular}{lcccc}
        \toprule
        Method & R1 & R2 & R3 & R4 \\
        \midrule
        Confidence
            & 0.550 & 0.530 & 0.558 & 0.520 \\
        CLIP Alignment
            & 0.538 & 0.551 & 0.549 & 0.527 \\
        Intervention
            & 0.511 & 0.493 & 0.532 & 0.465 \\
        Q-Former
            & 0.493 & 0.522 & 0.539 & 0.524 \\
        Binary MLP
            & 0.498 & 0.518 & 0.533 & 0.503 \\
        \bottomrule
    \end{tabular}
    \caption{Reason-wise answerability AUROC by unanswerability reason
    (R1--R4).}
    \label{tab:reason_auroc}
    \vspace{-8pt}
\end{table}

Reason-wise AUROC remains weak, with no method consistently reliable across
R1--R4. Intervention stays near chance for every reason, reinforcing that
visual dependence is not a reliable answerability signal.

\section{Discussion \& Limitations}
\label{sec:discussion}

\parhead{Visual dependence is not evidence sufficiency.}Targeted removal of question-relevant visual evidence produces a
substantially larger confidence drop than same-size random masking, confirming
that the frozen VLM is sensitive to localized visual evidence.
Yet intervention sensitivity remains near chance for answerability
discrimination.
Thus, detecting whether a model relies on visual evidence and determining
whether that evidence is sufficient to support a valid answer are distinct
problems.
This does not imply that visual evidence is irrelevant to answerability; rather,
visual dependence alone is insufficient as an answerability criterion.
A model may strongly rely on available visual content while still lacking other
information required to justify its answer.

\parhead{Frozen signals are weak standalone proxies for answerability.}Across uncertainty, alignment, grounding, visual dependence, and frozen
representations, answerability discrimination remains weak.
Feature fusion provides little improvement, while nonlinear probing improves
training separation without corresponding test gains, and reason-aware
supervision provides no consistent advantage.
These trends persist across bootstrap, multiple seeds, and post-freeze
cross-validation.
This does not imply that answerability information is absent from the VLM;
rather, it is not robustly recoverable from the static signals and probe
families evaluated here.

\parhead{Answerability failures are heterogeneous.}Relative performance changes across perturbations and reasons; unanswerability
is not a uniform state. Robust mechanisms may need to identify required
evidence and verify its availability rather than rely on one global score.

\parhead{Implications for selective answering.}Weak discrimination yields little feasible low-risk coverage, consistent with
evidence that calibration need not improve failure
separation~\cite{zhu2024revisiting}. Future selective answering should model
evidence sufficiency explicitly and abstain or request clarification when
required evidence is absent.

\parhead{Limitations.}We study one frozen backbone and a balanced UNK-VQA subset, so conclusions may
not transfer to architectures with different visual pathways
~\cite{qwen2025vl,zhu2025internvl3} or naturally occurring unanswerability.
Accordingly, our results should be interpreted as controlled comparisons of
proxy validity across balanced failure conditions, rather than as estimates of
performance under natural deployment prevalence.
The results cover only the tested frozen signals and probes. Although the
random-mask control shares masking distribution shift
~\cite{hesse2024benchmarking}, better evidence verification or jointly trained
abstention models may behave differently. Finally, the test subset was
inspected during development; frozen-method cross-validation supports the
trend, but a fully unseen benchmark would provide stronger evidence.

\section{Conclusion}
\label{sec:conclusion}

We systematically evaluated whether signals available from a frozen VLM can
reliably distinguish answerable from unanswerable visual questions.
Across predictive uncertainty, cross-modal alignment, visual grounding,
visual dependence, internal representations, and learned probes, answerability
discrimination remained consistently weak.
Increasing probe capacity, combining multiple answerability signals, and adding
reason-aware supervision did not provide robust improvements, and the same
overall pattern persisted across robustness and failure-mode analyses.

Most importantly, removing question-relevant visual evidence produced a
substantially larger confidence change than random removal, yet this
intervention signal remained near chance for answerability discrimination.
This distinction shows that \textit{visual dependence is not evidence
sufficiency}: a model may rely on relevant visual content without having
enough evidence to support a valid answer.
More broadly, our results expose a gap between signals that indicate model
reliance and signals that provide evidence for the evaluative claim of
answerability.
Our findings therefore motivate answerability methods that directly assess
whether the evidence required by a question is available and sufficient for
answering.

\clearpage
{
    \small
    \bibliographystyle{ieeenat_fullname}
    \bibliography{main}
}

\end{document}